\documentclass[runningheads]{llncs}

\usepackage{eccv}

\usepackage{eccvabbrv}

\usepackage{graphicx}
\usepackage{booktabs}
\usepackage{multirow}

\usepackage[accsupp]{axessibility}  

\usepackage{hyperref}

\usepackage{orcidlink}

\begin{document}

\title{Multimodal Ambivalence and Hesitancy Recognition via Cross-Attention and Gated Fusion} 

\titlerunning{Multimodal A/H Recognition}

\author{Oussama Berhili \and
Yassine Ouzar \orcidlink{0000-0002-7327-5895}\and
Larbi Boubchir \orcidlink{0000-0002-5668-6801}}

\authorrunning{Berhili et al.}

\institute{University of Paris 8, Saint-Denis, France \\
\email{oussama.berhili@etud.univ-paris8.fr ; \{yassine.ouzar, larbi.boubchir\}@univ-paris8.fr}}

\maketitle

\begin{abstract}
Recognizing ambivalence/hesitancy in unconstrained video settings remains a challenging task. These behavioral states are expressed through subtle, context-dependent cues that rarely manifest within a single modality, making them difficult to capture reliably. In this paper, we propose a multimodal system for video-level ambivalence/hesitancy recognition, developed for the ABAW11 challenge at ECCV 2026. The proposed approach relies on three modality-specific encoders: F2LLM-V2 for textual embeddings, WavLM-Large for acoustic features, and VideoMAE-V2 for modeling facial dynamics. The resulting unimodal embeddings are then integrated through a multimodal fusion strategy combining bidirectional cross-attention with a Gated Multimodal Unit (GMU). Experiments on the BAH dataset demonstrate the effectiveness of the proposed approach. The best unimodal configuration achieves a Macro F1-score of 0.6659, while our multimodal model reaches 0.7583, outperforming previous state-of-the-art methods and providing a relative improvement of 9.24\% over the strongest unimodal baseline. These results confirm that modeling cross-modal dependencies is essential for capturing the complex behavioral patterns underlying ambivalence and hesitancy. The source code is publicly available at: \url{https://github.com/yassineouzar/IUSD_AH/}.

\end{abstract}


\section{Introduction}
\label{sec:intro}
Human interaction is fundamentally nuanced. While major affective state like joy, anger, or sadness have been comprehensively mapped by affective computing frameworks, everyday decision-making and interpersonal communication are dominated by complex, blended states. 

Ambivalence and hesitation are among these phenomena. These two psychological states are interconnected and often constitute a major barriers to behavior change, leading individuals to postpone, avoid, or abandon decisions across multiple contexts, including health, education, professional life, financial decision-making, and social interactions \cite{gonzalez2025bah}. This often translates into decreased performance, missed opportunities, and inconsistent behavioral outcomes. Unlike discrete emotions such as happiness or sadness, ambivalence and hesitation reflect a subtle internal conflict between positive and negative valences. This conflict often manifests as a contradiction between what is said and how it is expressed, as well as through subtle and fleeting facial expressions that emerge throughout the course of speech. Such complexity makes the automatic recognition of ambivalence and hesitation particularly challenging, as it requires capturing weak, temporally dynamic, and often contradictory cues across multiple modalities, including facial behavior, vocal patterns, and linguistic content.

To support systematic study of these behavioral states, the Behavioural Ambivalence/Hesitancy (BAH) dataset and the corresponding Affective \& Behavior Analysis in-the-Wild (ABAW) challenge provide a standardized benchmark for A/H detection in naturalistic video interviews. The A/H challenge has evolved over three editions: the first focused on frame-level prediction, while the second reformulated the task as video-level classification, aiming to determine whether a given video exhibits A/H. The current edition maintains this video-level setting and introduces an expanded version of the BAH dataset with richer annotations, which serves as the basis for evaluating the present work.

Prior editions of the A/H challenge have led to a variety of proposed solutions, ranging from audiovisual systems addressing the
original frame-level formulation to more recent multimodal pipelines combining scene, facial, acoustic, and textual cues through
prototype-augmented or divergence-based fusion strategies for the video-level task~\cite{ryumina2026team,souza2026solution}. While they show promising results, they generally fail to effectively model fine-grained cross-modal interactions, limiting their ability to capture the complexity of ambivalence and hesitation.

In this paper, we propose a multimodal framework based on cross-attention and gated fusion. The cross-attention mechanism enables the model to explicitly capture inter-modal dependencies by dynamically aligning and weighting complementary cues across text, audio, and visual streams. In parallel, the gated fusion module adaptively regulates the contribution of each modality, allowing the model to emphasize the most informative modality while suppressing noisy or ambiguous ones. This design is particularly well-suited for A/H recognition, where subtle inconsistencies and modality-specific uncertainties play a central role. Experiments conducted on the BAH dataset demonstrate the effectiveness of the proposed approach in improving robust video-level A/H detection.

The contributions of this paper are summarized as follows :

\begin{itemize}
\item We provide a comprehensive evaluation of unimodal baselines for A/H recognition, comparing text, audio, and visual representations extracted from state-of-the-art pretrained encoders (F2LLM-v2-0.6B, WavLM-Large, and VideoMAE~V2) under three classical classifiers (MLP, Random Forest, GBDT).
\item We propose a multimodal fusion architecture that combines bidirectional cross-attention across modality pairs with a Gated
Multimodal Unit (GMU), allowing the model to both exchange fine-grained information between modalities and adaptively weight
each modality's contribution at the fusion stage.
\item We show that the proposed fusion strategy provides consistent improvements over all unimodal baselines and surpasses existing state-of-the-art methods on the BAH benchmark, demonstrating that explicit cross-modal interaction enables the model to capture complementary and potentially conflicting behavioral cues complementary behavioral cues beyond the capability of individual modalities.

\end{itemize}

\section{Related Work}
\label{sec:related}

The ABAW workshop series has driven progress on in-the-wild affective computing by releasing large-scale, naturalistic benchmarks covering valence-arousal estimation, action unit detection, and categorical emotion recognition \cite{kollias2026affect}. Earlier editions highlighted the challenges of in-the-wild data such as illumination, pose, occlusion, and speaker variability, motivating multimodal approaches beyond single-modality cues.

The BAH challenge extends prior work to encompass ambivalence and hesitancy as target behavioral states \cite{gonzalez2025bah}. A/H is qualitatively different from the discrete emotion categories typically studied in prior ABAW tracks, as it is defined less by a specific facial configuration than by the temporal and cross-modal coherence (or incoherence) of a response. Unlike discrete emotion categories typically studied in prior ABAW tracks, A/H is not characterized by a single modality or a specific pattern, but instead arises from the temporal dynamics and cross-modal consistency (or inconsistency) of multimodal behavioral signals such as speech, prosody, and visual cues.

Selecting suitable specific-encoder for each modality is crucial to extract rich, high-level representations from heterogeneous data sources. For the visual modality, most approaches rely on self-supervised video models such as VideoMAE and its variants \cite{tong2022videomae}, which effectively encode spatio-temporal dynamics from either full frames or aligned facial regions. Some methods instead rely on SigLIP2 \cite{tschannen2025siglip}, a vision-language pre-trained encoder that produces strong frame-level visual embeddings and offers an alternative to purely video-native backbones such as VideoMAE. For audio, WavLM-Large is widely adopted due to its strong ability to model speech characteristics from raw waveforms, generating high-dimensional temporal representations \cite{chen2022wavlm}. HuBERT \cite{hsu2021hubert} and Wav2Vec~2.0 \cite{baevski2020wav2vec} are common alternatives. For text, pretrained language models like F2LLM-V2 \cite{zhang2026f2llm} were used, which provide robust multilingual embeddings. Other approaches leverage models such as BERT \cite{devlin2019bert} or RoBERTa \cite{liu2019roberta}, while one study explores handcrafted psycholinguistic features to capture linguistic markers of uncertainty and hesitation.



Their findings revealed that text serves as the strongest unimodal predictor of A/H, yet achieving robust recognition performance demands specialised multimodal architectures and explicit temporal modelling capable of capturing the subtle cross-modal incongruities that characterise ambivalence and hesitancy.

 Their findings highlighted the strong contribution of textual cues at the unimodal level, while demonstrating that robust A/H recognition requires dedicated multimodal and temporal modeling to capture subtle cross-modal inconsistencies.

González-González et al. \cite{gonzalez2025bah} introduced a comprehensive benchmark encompassing supervised unimodal, bimodal, and trimodal settings based on facial, audio, and textual information, along with zero-shot and personalization scenarios. They systematically compared multiple fusion strategies, including naive concatenation, co-attention, transformer-based fusion, and cross-attention mechanisms. Their results show that text is the most informative single modality, but achieving reliable A/H recognition still requires dedicated multimodal and temporal modeling that can capture the subtle cross‑modal inconsistencies that characterize ambivalence and hesitation. Pereira et al. \cite{pereira2026brother} extracted semantic, acoustic, and visual embeddings with F2LLM \cite{zhang2025f2llm}, HuBERT \cite{hsu2021hubert}, and SigLip2 \cite{tschannen2025siglip}, respectively. To ensure that subtle temporal patterns were not diluted by these large global representations, they introduced a separate fourth modality capturing temporal statistical features. The final prediction was obtained through a Particle-Swarm-Optimization-based hard-voting ensemble that aggregates heterogeneous classifiers (MLP, Random Forest, and GBDT) trained on different modality combinations. Ryumina et al. \cite{ryumina2026team} built a four-modality system capturing scene, facial, acoustic, and textual cues. Scene dynamics were captured via VideoMAE \cite{tong2022videomae}, facial features were encoded through emotional frame-level embeddings aggregated by statistical pooling, acoustic representations were extracted with Emotion-Wav2Vec2.0 and processed by a Mamba-based temporal encoder, while textual content was handled by fine-tuned transformer-based text models. Bekhouche et al. introduced ConflictAwareAH \cite{bekhouche2026conflict}, a multimodal architecture specifically designed to exploit cross-modal discrepancies associated with A/H. Their framework leverages pretrained encoders, including VideoMAE, HuBERT, and RoBERTa-GoEmotions, to obtain visual, acoustic, and textual representations, which are aggregated through an attention-based pooling mechanism. The core innovation lies in computing pairwise conflict features as element-wise absolute differences between modality embeddings, thereby explicitly quantifying disagreement signals. This design directly addresses a critical limitation of text-dominant approaches by leveraging divergence as evidence of A/H and convergence as evidence of its absence.




\section{The BAH Dataset}
\label{sec:dataset}

The Behavioural Ambivalence/Hesitancy (BAH) dataset is a multimodal video corpus introduced to support automatic recognition of ambivalence and hesitation in digital behavioral change settings \cite{gonzalez2025bah}. It comprises 1{,}427 video recordings (approximately 10.6 hours in total, with an average of ~27 seconds per video) collected from 300 Canadian participants in naturalistic settings. During the recordings, participants respond to a set of seven questions designed to elicit a range of emotional and cogntive states, including ambivalence and hesitancy. Data collection was conducted via an online platform using an avatar-guided interaction protocol, aiming to approximate real-world personalized behavioral interventions. Each recording is annotated by behavioural‑science experts at both video and frame level with binary labels indicating the presence (1) or absence (0) of ambivalence/hesitation, and is accompanied by an automatic speech‑to‑text transcript and participant metadata. To avoid subject leakage, dataset partitions are defined at the participant level so that no individual appears in both training and evaluation splits.

The dataset is partitioned into training (778 videos), validation (124 videos), and test (525 videos) subsets. The training set is nearly balanced (393 negative vs. 385 positive samples), while the validation set contains 49 negative and 75 positive samples. Each recording includes the raw video, cropped and aligned facial frames at $256\times256$ pixels, audio at 16\,kHz, and time-stamped transcription files.

For the challenge, an additional unlabeled private test set—comprising 152 videos from 30 unseen participants was released separately. Participants were required to generate predictions without access to ground-truth labels, with a maximum of five submission attempts per team, either submitted jointly or incrementally.

\section{Proposed Approach}
\label{sec:method}


To capture a comprehensive behavioural representation for video-level A/H recognition, we extract features across three complementary modalities: textual (speech transcripts), acoustic (raw audio), and visual (aligned face frames). In contrast to approaches that collapse each modality into a single global embedding, we retain full token-level sequences during feature extraction, preserving fine-grained temporal and semantic information prior to fusion. Each modality is processed independently by a dedicated encoder (Sections~\ref{sec:text-extraction}, \ref{sec:audio-extraction}, and \ref{sec:visual-extraction}), producing modality-specific token embeddings. These sequences are subsequently integrated through a cross-attention and gated fusion module (Section~\ref{sec:fusion}), which explicitly models inter-modal dependencies at the video level while  leveraging the complementary strengths of each specialised encoder.

\subsection{Text encoding}
\label{sec:text-extraction}


The transcript of each video is tokenized with the official tokenizer of F2LLM-V2-0.6B \cite{zhang2026f2llm}, a decoder-only model built on Qwen3-0.6B and fine-tuned for text embedding
on 6 million open-source, non-synthetic examples. 

Given the importance of preserving subtle linguistic cues related to ambivalence and hesitancy, we avoid applying any pooling operation during feature extraction and retain the complete sequence of contextualized token embeddings. The resulting representation is a variable-length sequence $\mathbf{T} \in \mathbb{R}^{N_t \times 1024}$, where $N_t$ denotes the number of transcript tokens. This token-level representation is subsequently provided to the multimodal cross-attention module to enable fine-grained interactions with the acoustic and visual modalities.

\begin{figure}[!h]
    \centering
    \includegraphics[width=\columnwidth]{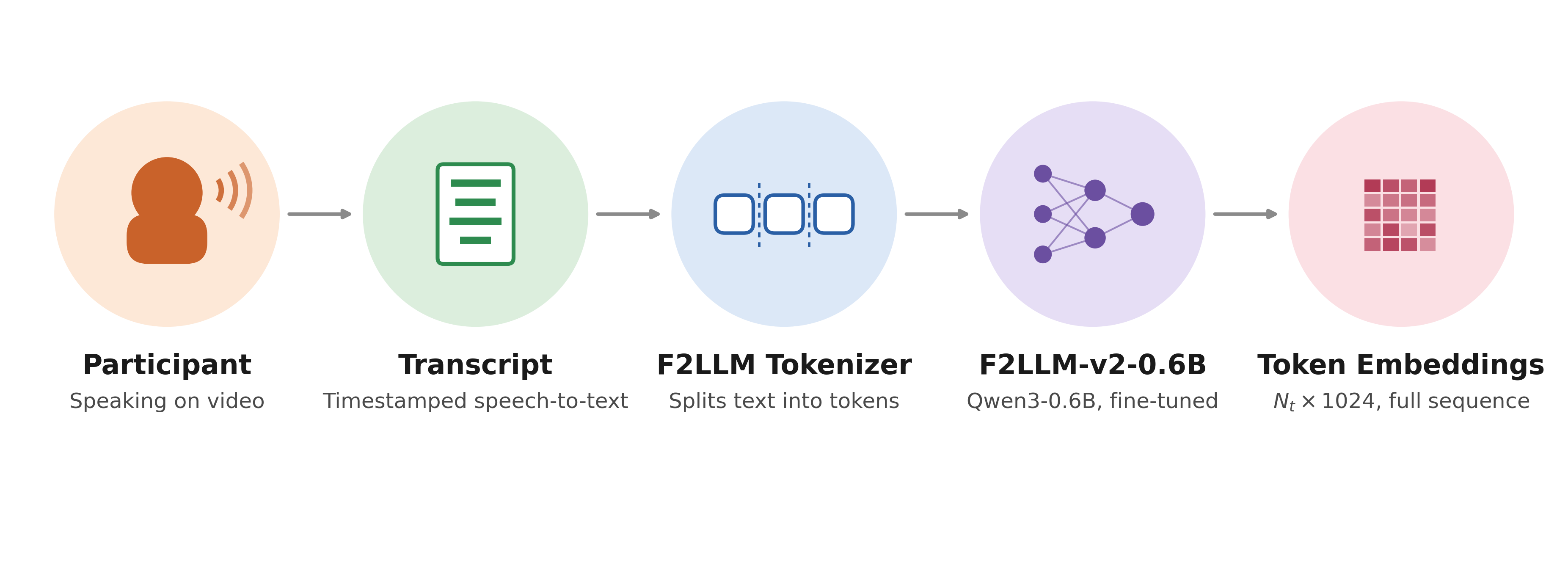}

        \caption{Overview of the textual encoding process. The transcript is tokenized using the F2LLM tokenizer and encoded with F2LLM-V2-0.6B to obtain the complete contextualized token sequence $\mathbf{T} \in \mathbb{R}^{N_t \times 1024}$, which is used for subsequent multimodal fusion.}

    \label{fig:text_extraction}
\end{figure}

\subsection{Audio encoding}
\label{sec:audio-extraction}

The audio stream is modeled directly from the speech signal associated with each video. After extracting the audio track with \texttt{ffmpeg} and standardizing it to a 16\,kHz mono waveform, we leverage WavLM-Large \cite{chen2022wavlm} to obtain contextualized speech representations. As a self-supervised speech foundation model, WavLM captures rich acoustic and temporal patterns from raw waveforms. Instead of applying temporal aggregation, we preserve the complete sequence of hidden representations, resulting in an acoustic feature matrix $\mathbf{A} \in \mathbb{R}^{N_a \times 1024}$, where $N_a$ corresponds to the number of audio tokens. This formulation retains fine-grained prosodic and temporal cues, such as pauses, intonation shifts, and hesitant speech dynamics, which provide valuable signals for A/H recognition.

\begin{figure}[!h]
    \centering
    \includegraphics[width=\columnwidth]{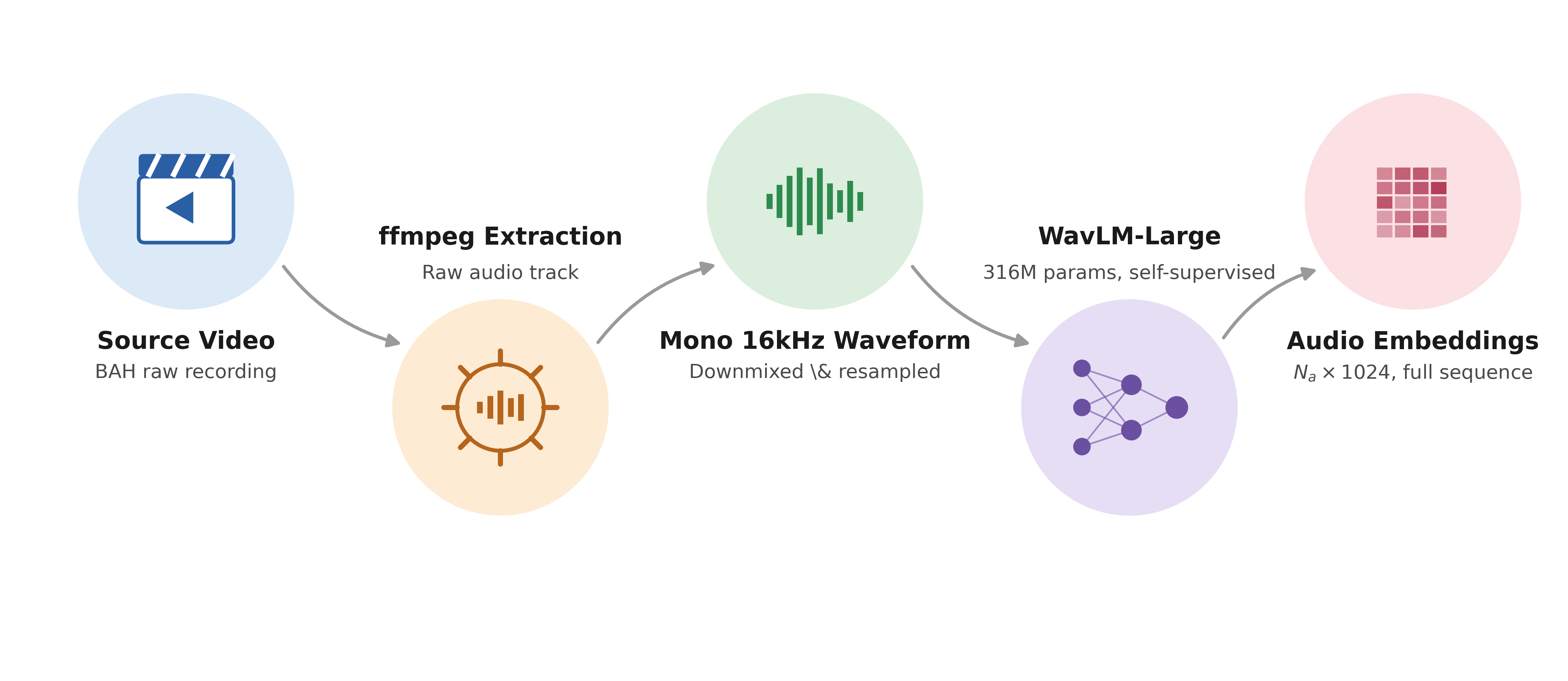}
    \caption{Audio feature extraction pipeline: raw audio is extracted from the source video via \texttt{ffmpeg}, downmixed to mono and resampled to 16\,kHz, then encoded by WavLM-Large. The full \texttt{last\_hidden\_state} sequence ($N_a \times 1024$) is preserved to retain fine-grained temporal information without applying pooling.}

    \label{fig:audio_extraction}
\end{figure}

\subsection{Visual encoding}
\label{sec:visual-extraction}

The visual modality is processed using VideoMAE-V2 \cite{tong2022videomae} applied to pre-extracted, aligned $256\times256$ facial frames. VideoMAE-V2 was selected over alternative video encoders due to its ability to jointly capture spatial appearance and temporal dynamics through spatio-temporal self-attention, rather than relying on independent frame-level representations followed by temporal aggregation. Since VideoMAE-V2 requires a fixed number of input frames, each video is split into non-overlapping chunks of 16 frames. For the last clip, when fewer than 16 frames remain, the sequence is completed by replicating the final frame to preserve all available information. Each clip is then processed by the VideoMAE-V2 transformer backbone, including patch embedding, positional encoding, transformer blocks, and final normalization, while avoiding CLS-token pooling to maintain token-level information. The resulting patch-level representations from all clips are concatenated, producing a variable-length visual representation $\mathbf{V} \in \mathbb{R}^{N_v \times 768}$, where $N_v$ denotes the total number of visual tokens. This formulation preserves fine-grained spatio-temporal patterns associated with facial dynamics and subtle behavioral cues throughout the entire video.

\begin{figure}[!h]
    \centering
    \includegraphics[width=\columnwidth]{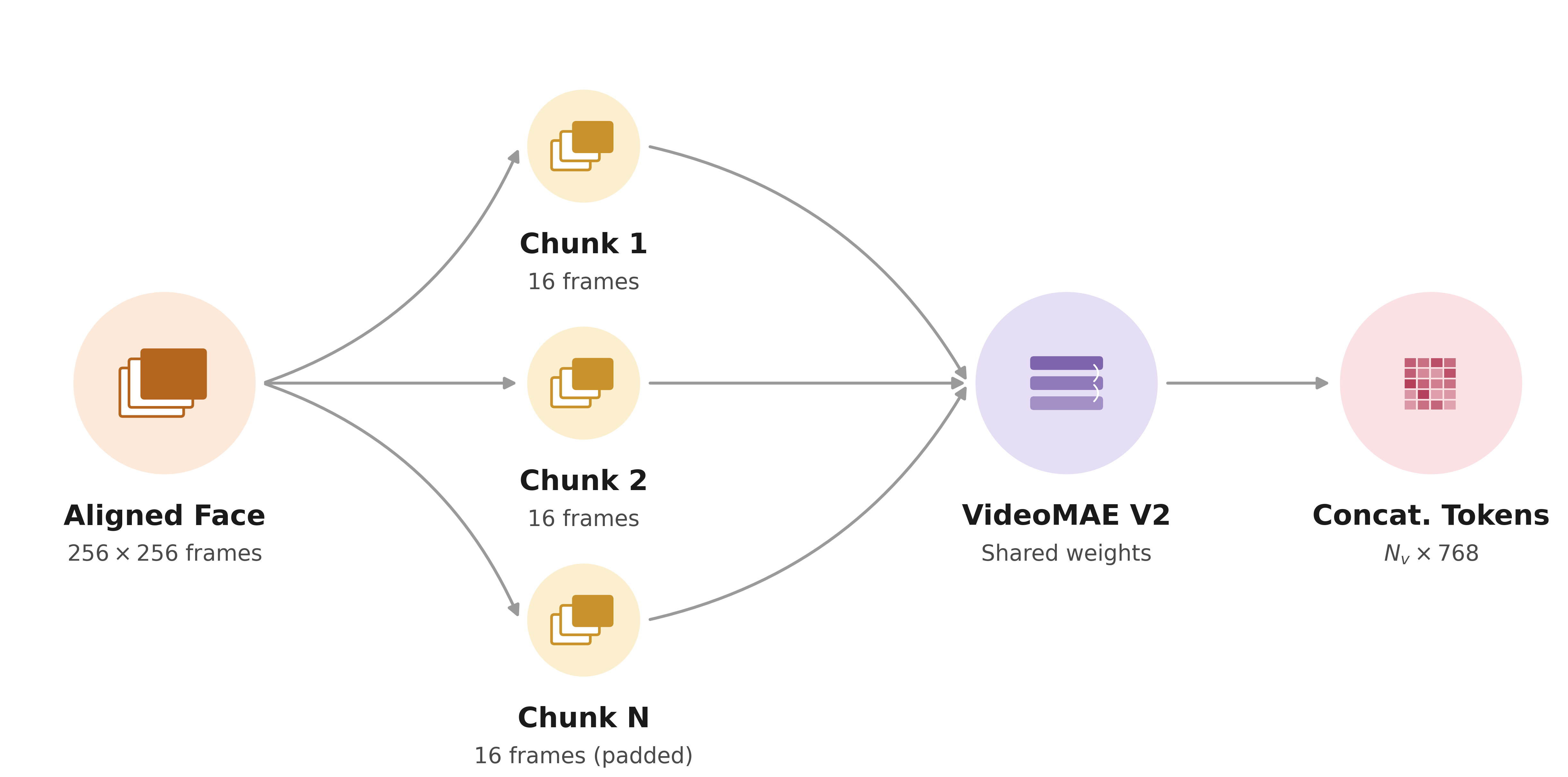}
    \caption{Visual feature extraction pipeline: aligned face frames are divided into fixed-length 16-frame clips and encoded independently using VideoMAE-V2. The patch-level token sequences extracted from all clips are then concatenated, preserving the complete spatio-temporal visual representation of the video ($N_v \times 768$).}
    
    \label{fig:visual_extraction}
\end{figure}

\subsection{Multimodal fusion}
\label{sec:fusion}

The extracted modality-specific representations contain complementary information about participant behavior, but they also exhibit heterogeneous temporal structures and modality-dependent uncertainties. Textual, acoustic, and visual signals do not contribute equally at every moment, and A/H-related cues often emerge from subtle agreements or inconsistencies across modalities. Therefore, instead of directly concatenating modality embeddings, we introduce a fusion strategy that explicitly models cross-modal interactions while adaptively controlling the contribution of each modality.

\begin{figure*}[t]
    \centering
    \includegraphics[width=0.98\textwidth]{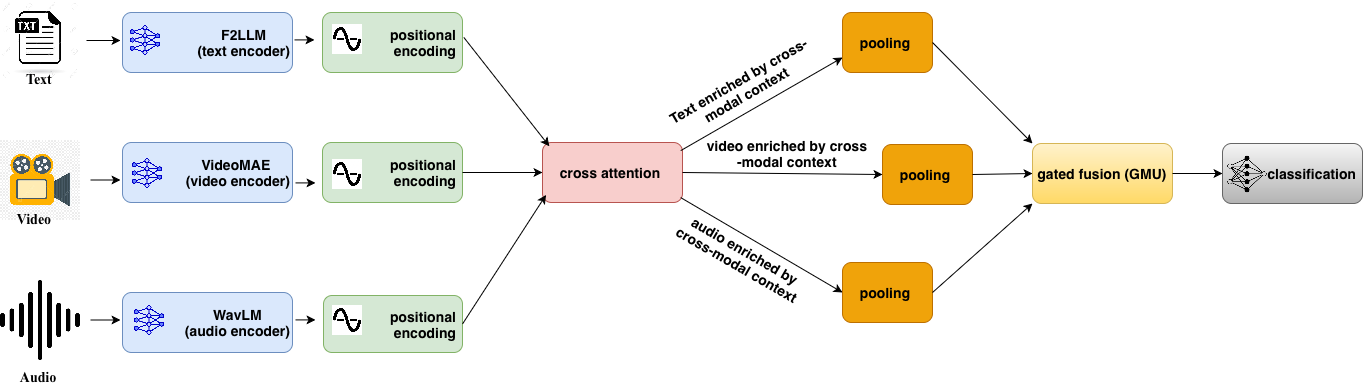}

    \caption{Overview of the proposed multimodal pipeline: each modality is first projected into a shared representation space with positional encoding, followed by bidirectional cross-attention to capture inter-modal dependencies. The resulting representations are aggregated and combined through a Gated Multimodal Unit (GMU), before final prediction using an MLP classifier.}
    \label{fig:pipeline_multimodal}
\end{figure*}

An overview of the proposed framework is presented in Figure~\ref{fig:pipeline_multimodal}, while the following sections detail the design and functionality of each component.

\subsubsection{Modality Projection and Positional Encoding}
\label{sec:proj-pe}

Prior to fusion, the audio and visual token sequences are uniformly resampled to fixed lengths ($N_a = 500$ and $N_v = 128$, respectively, see Section~\ref{sec:multimodal-results}), whereas textual representations preserve their original variable length $N_t$.

For each modality $m \in \{t,a,v\}$, with input sequence $X_m \in \mathbb{R}^{N_m \times d_m}$ (where $d_t=d_a=1024$ and $d_v=768$), a modality-specific linear projection is applied to map the representations into a shared embedding space of dimension $d_{\text{model}}$:

\begin{equation}
U_m = \mathrm{Dropout}\left(\mathrm{LayerNorm}\left(X_m W_m + b_m\right)\right)
\label{eq:projection}
\end{equation}

where $U_m \in \mathbb{R}^{N_m \times d_{\text{model}}}$ denotes the projected representation of modality $m$. The matrices $W_m \in \mathbb{R}^{d_m \times d_{\text{model}}}$ and $b_m \in \mathbb{R}^{d_{\text{model}}}$ correspond to the learnable modality-specific projection parameters. This operation aligns the heterogeneous modality representations into a common embedding space suitable for subsequent cross-modal interactions.

Since the cross-attention mechanism does not inherently encode the ordering of input tokens, we incorporate fixed sinusoidal positional embeddings~\cite{vaswani2017attention} into each projected sequence to preserve temporal information. For a position $pos \in \{0,\dots,N_m-1\}$ and embedding dimension index $i \in \{0,\dots,d_{\text{model}}/2-1\}$, the positional encoding is
defined as:
\begin{align}
PE_{(pos,\,2i)} &=
\sin\left(\frac{pos}{10000^{2i/d_{\text{model}}}}\right),
\label{eq:pe-sin}\\
PE_{(pos,\,2i+1)} &=
\cos\left(\frac{pos}{10000^{2i/d_{\text{model}}}}\right).
\label{eq:pe-cos}
\end{align}

The resulting modality representations, which serve as inputs to the cross-attention module, are obtained by:
\begin{equation}
Z_m^{(0)} = U_m + PE_m,
\qquad
Z_m^{(0)} \in \mathbb{R}^{N_m \times d_{\text{model}}}.
\label{eq:pe-add}
\end{equation}

\subsubsection{Cross-Modal Attention}
\label{sec:cross-attn}

Multimodal cross-attention module enables each modality to dynamically retrieve complementary information from the other modalities by alternating between query and context roles. Given a query sequence $Z_m$ and a context sequence $Z_c$ with padding mask $\mu_c \in \{0,1\}^{N_c}$ (where 1 indicates a valid token), we denote by $\mathrm{Attn}(Q,K,V,\mu)$ the multi-head attention operation with query $Q$, key/value pairs $K,V$, and a key-padding mask derived from $\mu$. Each cross-attention block is formulated as:

\begin{align}
\tilde{Z}_m &= \mathrm{LayerNorm}_q(Z_m), \\
\tilde{Z}_c &= \mathrm{LayerNorm}_{ctx}(Z_c), \\
A &= \mathrm{Attn}(\tilde{Z}_m,\,\tilde{Z}_c,\,Z_c,\,\mu_c),
\label{eq:mha} \\
Z_m' &= Z_m + \mathrm{Dropout}(A), \\
Z_m'' &= Z_m' +
\mathrm{Dropout}\left(
\mathrm{FFN}(\mathrm{LayerNorm}_{ffn}(Z_m'))
\right).
\label{eq:ffn-residual}
\end{align}

The position-wise feed-forward network is defined as:
\[
\mathrm{FFN}(z)=
W_2\,\mathrm{Dropout}
\left(
\mathrm{GELU}(W_1z+b_1)
\right)+b_2,
\]
where $W_1 \in \mathbb{R}^{d_{\mathrm{model}}\times d_{\mathrm{ffn}}}$ and
$W_2 \in \mathbb{R}^{d_{\mathrm{ffn}}\times d_{\mathrm{model}}}$ are learnable
parameters. For variable-length sequences, particularly textual
representations, padded positions ($\mu_c=0$) are masked out during attention
computation and therefore do not contribute to the attended context in
Eq.~\eqref{eq:mha}.

To capture bidirectional interactions among the three modalities, we employ six
directional cross-attention pathways:
{text$\leftarrow$video}, text$\leftarrow$audio,
{video$\leftarrow$text}, video$\leftarrow$audio,
{audio$\leftarrow$text}, and audio$\leftarrow$video.
Each pathway consists of one or more stacked cross-attention blocks. The
modality representations are progressively refined by attending to
complementary streams, enabling the model to capture complex cross-modal
dependencies and resolve ambiguous behavioral patterns. For instance, hesitant
vocal cues can be interpreted more accurately when jointly considered with
supporting visual and linguistic evidence.

\subsubsection{Pooling}
\label{sec:pooling}

The Gated Multimodal Unit described below operates on a single fixed-dimensional vector per modality, whereas the output of the cross-attention stage remains a token-level sequence ($Z_m'' \in \mathbb{R}^{N_m \times d_{\mathrm{model}}}$). Therefore, a pooling operation is introduced as an intermediate aggregation step to transform each contextualized modality sequence into a compact
representation. This aggregation summarizes the information captured through both intra-modal and cross-modal interactions while removing the temporal/token dimension prior to multimodal fusion.

We investigate three alternative pooling strategies: mean pooling, max pooling, and learnable attention pooling. Each strategy is followed by layer normalization to improve representation stability. The pooling mechanism is treated as an architectural hyperparameter and evaluated in Section~\ref{sec:multimodal-results}, as different aggregation functions emphasize different characteristics of the sequence, ranging from global behavioral patterns to highly discriminative temporal events.

\subsubsection{Gated Multimodal Fusion}

The three pooled modality representations are concatenated and fed into the
Gated Multimodal Unit (GMU). The GMU first derives a joint context representation
from the aggregated features, which is then used to estimate a
modality-specific sigmoid gate. Each modality representation is weighted by its
corresponding gate before being combined through a weighted summation. This
adaptive mechanism allows the model to dynamically adjust the contribution of
each modality, attenuating unreliable or noisy signals (e.g., degraded visual
information or low-quality audio) while emphasizing the most informative cues
for each sample.


\subsubsection{Classification}
\label{sec:classification}

The final fused representation $f \in \mathbb{R}^{d_{\mathrm{model}}}$ is
forwarded to a multi-layer perceptron (MLP) classification head. The classifier
architecture is configurable in terms of the number of layers, hidden
dimensions, activation function, and dropout rate, and produces a single output
logit:
\begin{equation}
\hat{y} = \mathrm{MLP}(f) \in \mathbb{R},
\label{eq:classifier}
\end{equation}

where the MLP applies an initial layer normalization to $f$, followed by a
sequence of linear-activation-dropout blocks and a final linear projection to
obtain the classification logit. The entire network is optimized end to end
using the binary cross-entropy loss computed from the predicted probability and
the ground-truth A/H label $y \in \{0,1\}$:
\begin{equation}
\mathcal{L} =
-\left[
y\log\sigma(\hat{y})+
(1-y)\log\left(1-\sigma(\hat{y})\right)
\right],
\label{eq:bce}
\end{equation}

where $\sigma(\cdot)$ denotes the sigmoid activation function. During inference,
the probability of A/H presence is obtained as $\sigma(\hat{y})$, and a threshold
of 0.5 is applied to convert the probability score into the final binary
prediction.


\section{Experiments}

\subsection{Evaluation Metric}
 
The A/H recognition task is formulated as a binary video-level classification problem, where the
objective is to determine whether a given video exhibits A/H. Since the dataset presents class imbalance, we adopt Macro F1-score as the primary evaluation metric, following the official A/H challenge protocol. Macro F1 gives equal importance to both classes by averaging the F1-score
computed independently for the positive and negative classes. 

The F1-score is defined as:

\begin{equation}
F1 = \frac{2 \times Precision \times Recall}
{Precision + Recall},
\end{equation}

where precision and recall are computed from the predicted binary labels. The Macro F1-score is obtained by averaging the F1 values of the two classes.

\subsection{Hyperparameter Optimization}
\label{sec:optimization}

To ensure a fair comparison between different models, all trainable classifiers
and architectural components are optimized using the same validation-based
protocol. Hyperparameter optimization is performed with Optuna~\cite{optuna2019},
using a Tree-structured Parzen Estimator (TPE) sampler over 100 trials for each classifier. The optimization objective is to maximize the Macro F1-score on the validation set. The search space includes model-specific parameters, such as hidden-layer dimensions, number of layers, and regularization factors for MLP, as well as the number of estimators, tree depth, and learning-related parameters for RF and GBDT. The final classifiers are selected based on the configurations achieving the highest validation Macro F1-score and are then evaluated on the held-out test set.

\subsection{Unimodal Evaluation}
\label{sec:unimodal}
Before evaluating the proposed multimodal fusion strategy, we first analyze the contribution of each individual modality. Textual, acoustic, and visual representations are evaluated separately using three supervised classifiers: Multi-Layer Perceptron (MLP), Random Forest (RF), and Gradient Boosted Decision Trees (GBDT). Each classifier is trained exclusively on a single modality, providing a controlled assessment of the discriminative power of individual behavioral cues and their limitations for A/H recognition.

Table~\ref{tab:unimodal} reports the performance of our supervised unimodal models, together with the zero-shot baseline provided by the BAH challenge organizers. This baseline, obtained
using Video-LLaVA \cite{lin2024video} without task-specific fine-tuning, serves as an external reference to assess the benefit of adapting models to the A/H recognition task.

All supervised approaches substantially outperform the challenge-provided
zero-shot baseline across the three modalities, demonstrating the importance of
task-specific training for this problem. This performance gap highlights that
general-purpose multimodal models, while powerful for broad video understanding
tasks, struggle to capture the subtle behavioral patterns required for
ambivalence and hesitation recognition without adaptation to the target
domain. Among the three modalities, textual representations achieve the strongest
performance, with a best Macro F1-score of 0.6659 obtained using the Random
Forest classifier. Audio representations follow with a Macro F1-score of 0.6275,
while visual features reach 0.5727 with the MLP classifier. The superiority of
the textual modality indicates that linguistic content provides highly
informative cues for A/H detection, as hesitancy and ambivalence are often
expressed through uncertainty markers, cautious formulations, and conflicting
statements. However, the lower performance of the acoustic and visual streams
does not imply that these modalities are irrelevant; rather, their cues are
typically more implicit and context-dependent, requiring temporal modeling and
cross-modal reasoning to be effectively exploited.

The classifier comparison also reveals modality-dependent behaviors. Tree-based
models perform particularly well on textual and acoustic embeddings, with
Random Forest achieving the best results for both modalities, suggesting that
nonlinear decision boundaries can effectively capture discriminative patterns
in these representations. In contrast, the visual modality benefits more from
the MLP classifier, indicating that the high-dimensional visual embeddings may
require a learned nonlinear projection rather than direct partitioning by tree
structures.

Overall, these unimodal results confirm that no single modality fully captures
the complexity of A/H. While textual cues provide the strongest individual
signal, acoustic and visual modalities contain complementary behavioral
information. This motivates the use of a dedicated multimodal fusion strategy
capable of modeling interactions between modalities rather than treating them
as independent sources.

\begin{table}[t]
\centering
\caption{Unimodal classification results on the BAH dataset.}
\label{tab:unimodal}
\small
\setlength{\tabcolsep}{5pt}
\renewcommand{\arraystretch}{0.9}
\begin{tabular}{llccc}
\toprule
\textbf{Modality} & \textbf{Classifier} & \textbf{Macro F1} & F1$_0$ & F1$_1$ \\
\midrule
\multirow{3}{*}{Text}
& MLP  & 0.6550 & 0.6096 & 0.7003 \\
& RF   & \textbf{0.6659} & 0.6360 & 0.6958 \\
& GBDT & 0.6505 & 0.5959 & 0.7051 \\
\midrule
\multirow{3}{*}{Audio}
& MLP  & 0.5946 & 0.5336 & 0.6556 \\
& RF   & \textbf{0.6275} & 0.5494 & 0.7055 \\
& GBDT & 0.5717 & 0.5089 & 0.6346 \\
\midrule
\multirow{3}{*}{Visual}
& MLP  & \textbf{0.5727} & 0.5206 & 0.6248 \\
& RF   & 0.5364 & 0.4905 & 0.5823 \\
& GBDT & 0.5368 & 0.4926 & 0.5809 \\
\midrule
Zero-shot & Video-LLaVA & 0.2827 & -- & -- \\
\bottomrule
\end{tabular}
\end{table}

\subsection{Multimodal Evaluation}
\label{sec:multimodal-results}

Following the unimodal analysis, we evaluate whether jointly modeling textual,
acoustic, and visual information improves A/H recognition. The proposed
multimodal framework combines the three modality-specific token representations
through bidirectional cross-attention, which enables each modality to retrieve
complementary information from the others. The resulting contextualized
representations are then aggregated and integrated using a Gated Multimodal
Unit, which dynamically adjusts the contribution of each modality based
on its relevance for the current sample. 

Table~\ref{tab:multimodal} reports the comparison with recent methods evaluated
on the BAH benchmark. The proposed Cross-Attention + GMU architecture
achieves a Macro F1-score of 0.7583, obtaining the best performance among all
compared approaches. It surpasses ConflictAwareAH \cite{bekhouche2026conflict}
(0.694), LEYA \cite{ryumina2026team} (0.7143), AffectGPT \cite{gonzalez2026multimodal}
(0.7229), and BROTHER~\cite{pereira2026brother} (0.7465), including a gain of 1.18 percentage points over the previous best-performing method.

These results highlight the importance of explicitly modeling interactions between modalities for A/H recognition. While previous approaches have explored ensemble-based fusion, modality-specific weighting, or conflict modeling, the proposed method enables fine-grained information exchange through cross-attention before fusion and dynamically adjusts modality contributions through GMU. This is particularly relevant for A/H, where the target behavioral state may emerge from subtle temporal relationships and potential inconsistencies between linguistic, acoustic, and visual signals.

Overall, the performance improvement demonstrates that adaptive cross-modal
reasoning provides a more effective strategy than independent modality modeling
or fixed fusion mechanisms for capturing the complex behavioral patterns
associated with ambivalence and hesitation.

\begin{table}[t]
\centering
\caption{Multimodal classification results on the BAH dataset.}
\label{tab:multimodal}
\setlength{\tabcolsep}{4pt}
\renewcommand{\arraystretch}{0.9}
\small
\begin{tabular}{lc}
\toprule
\textbf{Method} & \textbf{Macro F1} \\
\midrule
ConflictAwareAH \cite{bekhouche2026conflict}  & 0.694 \\
LEYA \cite{ryumina2026team} &  71.43 \\
AffectGPT \cite{gonzalez2026multimodal}  & 0.7229 \\
BROTHER \cite{pereira2026brother} & 0.7465 \\
\textbf{Cross-Attention + GMU (Ours)} & \textbf{0.7583} \\
\bottomrule
\end{tabular}
\end{table}

\section{Conclusion}
\label{sec:conclusion}
In this work, we investigated the recognition of Ambivalence and Hesitancy
in videos through a systematic evaluation of unimodal and multimodal learning
strategies on the ABAW11 BAH benchmark. We first established modality-specific
baselines by leveraging state-of-the-art pretrained encoders for text, audio,
and visual signals, combined with optimized classical classifiers. These
experiments revealed that textual information provides the strongest individual
signal, while acoustic and visual modalities offer complementary behavioral
cues that cannot be fully captured in isolation.

To exploit these complementary signals, we introduced a multimodal architecture
based on bidirectional cross-attention and Gated Multimodal Unit fusion.
By explicitly modeling interactions between modalities and adaptively weighting
their contributions, the proposed approach achieved a Macro F1-score of 0.7394
on the BAH test set, outperforming unimodal alternatives and demonstrating
the importance of cross-modal reasoning for A/H recognition. These results
highlight that ambivalence and hesitancy emerge from complex behavioral
patterns distributed across multiple modalities rather than from a single
observable cue.

The proposed model was further used to generate predictions on the official
private test set following the ABAW challenge protocol. Future research
directions include improving temporal representation learning, exploring model
ensembling strategies, incorporating personalization and domain adaptation
techniques, and investigating parameter-efficient adaptation of multimodal
foundation models for behavioral state recognition.

%
%

\bibliographystyle{unsrt} 
\bibliography{main}

\end{document}